\documentclass[runningheads]{llncs}
\usepackage{bbding}
\usepackage{graphicx}
\usepackage{amsmath,amssymb,amsfonts}
\usepackage{algorithm} 
\usepackage{algpseudocode} 
\usepackage[export]{adjustbox}
\usepackage{xcolor}
\usepackage{wrapfig}
\begin{document}
%

\title{Adversarial Robustness in Deep Learning: Attacks on Fragile Neurons
}
\titlerunning{Adversarial Robustness in Deep Learning}
%
\author{\Envelope~Chandresh Pravin\inst{1}\orcidID{0000-0003-1530-0121}  \and
Ivan Martino\inst{2}\orcidID{0000-0001-6306-6777}  \and
Giuseppe Nicosia\inst{3,4}\orcidID{0000-0002-0650-3157}  \and
Varun Ojha\inst{1}\orcidID{0000-0002-9256-1192}}
\authorrunning{C. Pravin et al.}
%
\institute{
University of Reading, UK \\
\email{\{kp826252,v.k.ojha\}@reading.ac.uk}
\and
KTH Royal Institute of Technology, Sweden\\
\email{imartino@kth.se}
\and
University of Catania, Italy\\
\and
University of Cambridge, UK\\
\email{giuseppe.nicosia@unict.it,gn263@cam.ac.uk}
}
\maketitle              
%
\begin{abstract}
We identify fragile and robust neurons of deep learning architectures using nodal dropouts of the first convolutional layer. Using an adversarial targeting algorithm, we correlate these neurons with the distribution of adversarial attacks on the network. Adversarial robustness of neural networks has gained significant attention in recent times and highlights an intrinsic weaknesses of deep learning networks against carefully constructed distortion applied to input images. In this paper, we evaluate the robustness of state-of-the-art image classification models trained on the MNIST and CIFAR10 datasets against the fast gradient sign method attack, a simple yet effective method of deceiving neural networks. Our method identifies the specific neurons of a network  that are most affected by the adversarial attack being applied. We, therefore, propose to make fragile neurons more robust against these attacks by compressing features within robust neurons and amplifying the fragile neurons proportionally.   

\end{abstract}
\section{Introduction}
\label{sec:inro}
Deep neural networks (DNNs) have been widely adapted to various tasks and domains, achieving significant performances in both the real world and in numerous research environments~\cite{lecun2015deep}. Previously considered state-of-the-art DNNs have been subjected to a plethora of tests and experiments in an attempt to better understand the underlying mechanics of how and what exactly these learning models actually learn~\cite{papernot2016limitations}. In doing so, we now better recognise the strengths and more importantly the weaknesses of DNNs and have subsequently developed better networks building on from previous architectures~\cite{silva2020opportunities}.

Adversarial attacks are one the most used methods to evaluate the robustness of DNNs. Such methods introduce a small carefully crafted distortion to the input of the network in an attempt to deceive the network into misclassifying the input with a high level of confidence~\cite{goodfellow2015explaining,szegedy2014intriguing}.
The small distortions to the input, termed \textit{adversarial perturbations}, are hardly perceptible to humans, even when the perturbation is amplified by several orders of magnitude~\cite{szegedy2014intriguing}. This ability to fool DNNs with hardly perceptible changes in the input highlights an intrinsic difference between artificial intelligence and true intelligence.  

There are many ways in which an adversarial perturbation can be crafted, utilising various tools and assumptions on the target model and dataset. Existing adversarial attacks, and methods for designing such distortions, can be broadly categorised into white-box and black-box attacks. The distinction between the two different types of attacks being the information that the adversary has on the model and its parameters. With the white-box attacks, the adversary is assumed to have complete access to the target model in question, including model parameters and architecture~\cite{carlini2017magnet}. Conversely, the black-box attack is a type of perturbation designed by an adversary with no information to the model's parameters or architecture~\cite{REN2020346}. In this paper, we focus our efforts at evaluating the robustness~\cite{Nicosia11} of ResNet-18, ResNet-50 and ResNet-101 networks against a simple yet effective white-box adversarial attack, the fast gradient sign method (FGSM) attack~\cite{goodfellow2015explaining}. We apply the FGSM perturbations on the MNIST and CIFAR10 datasets for the mentioned models and present a correlative relationship between the distribution of neurons with high influence and targeting by an adversary. We also evaluate a method in minimising the effects of such distortions.

With the numerous adversarial attacks formed against DNNs, there have been equally as many defences proposed in literature~\cite{REN2020346}. The ability of a defence model to remain unbeaten by an ever-growing selection of adversarial attacks has proven to be difficult~\cite{REN2020346,yuan2019adversarial}. Adversarial defences, much like adversarial attacks, can be divided into different categories: (i) defences focusing on gradient masking/obfuscation, whereby the network weight gradients used by adversaries to form attacks are disguised; (ii) robust optimization~\cite{Nicosia11b}, where the network structure/parameters are altered to increase adversarial defences; and (iii) adversarial example detection, where the goal is to detect an adversarial input and process this entity differently to ordinary inputs~\cite{silva2020opportunities}.

The goal of all adversaries is to deceive the network into predicting, classifying, or recognising an input as a different class to its true self. When the adversary has knowledge of the information held by the network, as is the case for white-box attacks, it utilises this to craft a perturbation that will exploit weaknesses within the network's representations of the data~\cite{REN2020346}. In this paper, we propose viewing an adversarial attack as an exploitative method that targets specific neurons within a given layer. We also draw a relationship between the adversaries' target neurons and neurons that show to have higher influence on the model's unperturbed performance.

We assume that, for a given layer, information about the input learned by the layer through back propagation is distributed unevenly amongst individual neurons. We propose using \textit{nodal dropout} to find redundant nodes within a given layer of a network~\cite{li2018first}. Thus, also finding \textit{fragile neurons} that carry more information about the input~\cite{goh2021multimodal}.  
We identify \textit{null neurons} that once removed do not significantly affect the overall model performance and thus considered to carry less information about the dataset. 
We examine how the FGSM attack affects different models (ResNet-18, ResNet-50 and ResNet-101) at different stages (epochs) in learning, whilst also comparing how increasing the network architecture affects the effectiveness of the formed attack. 
Therefore, we propose to make fragile neurons more robust against these attacks by compressing robust neurons and amplifying the fragile neurons proportionally.

Furthermore, the FGSM attack utilises a given network's learned representations in the form of its layer weights to calculate an effective adversarial example. The adversarial examples can be used as a method of evaluating the robustness~\cite{NicosiaN11t} of the model's composite representations. We aim to identify the fragile and robust neurons within specific parts of the network, and post-process them separately to investigate how they affect the overall model's robustness against an adversarial attack.

\section{Related Work} 
\label{sec:related_work}
Robustness analysis evaluates the defence of DNNs against malicious distortion of its input~\cite{akhtar2018threat,goodfellow2015explaining,yuan2019adversarial}. There are different types of attacks available for a potential adversary, each with their own strengths and limitations. Szegdey et al.~\cite{szegedy2014intriguing} initially proposed adversarial examples for DNNs using the \textit{Limited-memory Broyden-Fletcher-Goldfarb-Shanno} (L-BFGS) algorithm, an expensive linear search method for adversarial examples. Thereafter, the FGSM attack proposed by Goodfellow et al.~\cite{goodfellow2015explaining} has become one of the benchmarks for adversarial attacks due to its computational process being less resource intensive when compared to other attacks. 
The FGSM attack performs a pixel-wise one step gradient update along the gradient sign direction of increasing loss. 
%
%
%
There are several other attack methods available in the literature. However, in this study, we focus specifically on the FGSM adversarial attack due to its one-step gradient calculation and effective performance against state-of-the-art DNN models.

In terms of defences against adversarial attacks, there are an equal number of approaches proposed in literature. For every newly developed adversarial attack, soon there have been suitable defences proposed by researchers~\cite{yuan2019adversarial}.
One method of defending a DNN model is by masking the network's parameters, therefore making it more difficult for an adversary to exploit the network's learned information to generate adversarial examples. However, this method has shown to be ineffective against many types of attacks and there exist techniques to circumvent such defensive measures~\cite{silva2020opportunities}. 
Some studies show that adversarial examples are drawn from a different distribution to the regular dataset~\cite{grosse2017statistical}. Therefore, one method to defend against the effects of such adversarial examples is to identify them and deal with the perturbed inputs to the model separately~\cite{silva2020opportunities}. These methods are also subjected to exploitation by techniques that can bypass the adversarial examples detection, making such defence methods weaker to some types of attacks~\cite{carlini2017adversarial}.


In this paper we try to find a relationship between highly influential neurons and the likelihood of being targeted by an adversary, and accordingly, propose a method of regularising the specific neurons during post-training. As we, the observer, propagate through the network we notice that the deconstructed, abstract characteristics of the data begin to take a shape of salient features, which are then assigned semantic meaning in the form of target labels~\cite{zhou2018interpreting}. Literature on leveraging the information content of a DNN has been used for various applications, we direct the reader to Golatkar et al.~\cite{golatkar2020eternal} and their method of selective forgetting, in which they propose a framework for erasing the information about a particular subset of data from the model's learned weights. We take inspiration from this framework and propose that adversarial robustness is hinged on the distribution of \textit{influential} and \textit{uninfluential} neurons, referred to as sets $S$ and $S'$ respectively within the context of this study.

We are motivated by the works of Li and Chen~\cite{li2018first} along with related literature in reducing network complexity by using techniques such as nodal pruning. We leverage the idea that there exist neurons within a network that can be classified as redundant, or uninfluential to the overall model performance. Removing redundant neurons in some cases also shows to improve robustness against attacks~\cite{cheney2017robustness}. Conversely, we also consider the works of~\cite{goh2021multimodal} that prove the existence of multi-model neurons within networks; multi-modal neurons being representations that hold a higher degree of influence in the network's understanding of the data. We investigate the correlation between representations that show a higher influence and the highest average concentration of an adversarial attack to these features. In consequence, we draw attention to the nature of adversarial attacks and how such perturbations target the model's learned knowledge specifically.

\section{Adversarial Attack and Defense Formulations}
\label{sec:problem}
We consider an image classifier model $f_\theta$ with $L$ layers, and trainable parameters $\theta$ that accepts an input image $x$ and its associated true class label $y$. The model returns $\hat{y}$ as its prediction for input $x$. The goal of the model is to reduce loss function $\mathcal{L}(f_\theta, x, y)$. 
The image $x' = x + \delta_\epsilon$ is an adversarial example produced by a distortion $\delta_\epsilon$ added to image $x$, where $\epsilon$ is the perturbation magnitude.

Our objective is to minimise the difference in predictions values $\hat{y}$ obtained for unperturbed input $x$ and perturbed input $x'$.  
We examine the model's learnable parameters $\theta_L \in \theta$ of layer $L$ at various stages of the model's training. It should be noted that while assessing the significance of the neurons, we remove one-neuron at a time from  $\theta_L$. We, therefore, identified two sets of neurons indices, $S$ and $S'$ respectively representing (i) neuron indices within the layer $L$ showing a higher influence on the overall model performance, and (ii) neurons indices with lower overall influence on model performance. 
In our work, we are concerned with removing one-neuron at a time, removing multiple neurons from the model $f_\theta$  would warrant an alternative method. We also assessed neurons of the first layer of the network because of its high importance and influence on features learned by subsequent layer in a network~\cite{cheney2017robustness}.

\subsection{Attack Formulation}
We formulated attack in this work using FGSM method. This method leverages a network's learned representations in the form of layer weights $\theta_L$ to construct an efficient and effective adversarial perturbation $x'$. The FGSM attack is a perturbation for an input $x$ computed as:
\begin{equation}
    \delta_{\epsilon} = \epsilon \textit{ sign} (\nabla_x \mathcal{L}(x,y,\theta)),
\end{equation}
where $\nabla_x$ is the required gradients calculated using backpropagation. The adversarial example therefore is $x' = x + \delta_\epsilon$~\cite{akhtar2018threat,goodfellow2015explaining}. 

We find that for a 100 epoch pre-trained ResNet-50 model on the CIFAR10 dataset, a baseline model accuracy of $75.87\%$ on unperturbed input $x$ is found. The same model applied to the CIFAR10 dataset with an FGSM attack, using a perturbation magnitude of $\epsilon=0.01$, results in an accuracy of $58.88\%$. If we consider the same ResNet-50 architecture trained equally for 100 epochs, with the input dimensions adjusted to comply with the MNIST dataset, the baseline model accuracy on unperturbed MNIST dataset is $99.42\%$. While the model accuracy is found to be $79.4\%$ when perturbed with an $\epsilon=0.34$ attack. These are examples of the FGSM attack performance against CIFAR10 and MNIST datasets on the ResNet-50 DNN model. 

If we consider a metric to assess the complexity of a given dataset, such as the cumulative spectral gradient (CSG) method~\cite{Branchaud-Charron_2019_CVPR}, we notice that the CSG complexity measure for the for the CIFAR10 dataset is $1.00$ and MNIST dataset is $0.11$. As we may expect, the FGSM attack is more effective on more complex data (e.g., CIFAR10) compared to less complex data (e.g., MNIST). This can be realised from the perturbation magnitude $\epsilon$ required for the model performance to decrease proportionally. For example, to decrease performance by approximately $20\%$, a lower value of $\epsilon$ (small perturbation) is required for CIFAR10 and a higher value of $\epsilon$ (large perturbation) is required for MNIST.

\subsection{Defence Formulation}
%
%
To better understand how to form a suitable defence against an adversarial attack, we may consider how an adversary can form an effective attack. With the FGSM attack, a single step in the parameter space is taken in the direction of increasing loss. The perturbation is calculated using the network's weights to perturb the input data features in the direction of an incorrect class. 
Then it is natural to consider that this informed way of creating adversarial perturbations may, even with relatively low magnitudes, affect the neurons that are more influential to the model's performance (e.g., set of highly influencing neurons $S$). 

We aim to show this effect of adversarial perturbations experimentally by comparing the output of the layer-wise convolution for original input $x$ and perturbed input $x'$ computed using pre-trained parameters $\theta$. We expect the original model prediction $f(x,y,\theta)$ and the model prediction on perturbed input $f(x',y,\theta)$ to be not equal. In our defence formulation, we aim to modify the model's layer parameters $\theta_L$ as $\theta'_L$ such that a potential adversary is forced to distribute the attack strength throughout the layer. We propose that this will make the model's layer $\theta'_L$ more robust against an adversarial attack.

\subsection{Fragile and Null Kernels Identification}
\label{sec:frigile_and_null_kernerls}
We identified fragile neurons (kernels) $S$ and null neurons (kernels) $S'$ by dropping the kernels out systematically one-by-one and measuring the variance in model performance. Fig.~\ref{fig:kernal_droput_exp} show model's performance for each kernel along the x-axis being dropped. The indices of fragile kernels $S$ are indicated with blue circled symbols and are below the mean performance line indicated in red, which is computed over each kernel's effect on the model's accuracy. The dropping of these fragile kernels has a higher influence on the model's performance when compared to the dropping of the null kernels indicted with black star symbol shown above mean performance line.

{\setlength\intextsep{15pt}
\begin{figure}
    \centering
    \includegraphics[width=\textwidth]{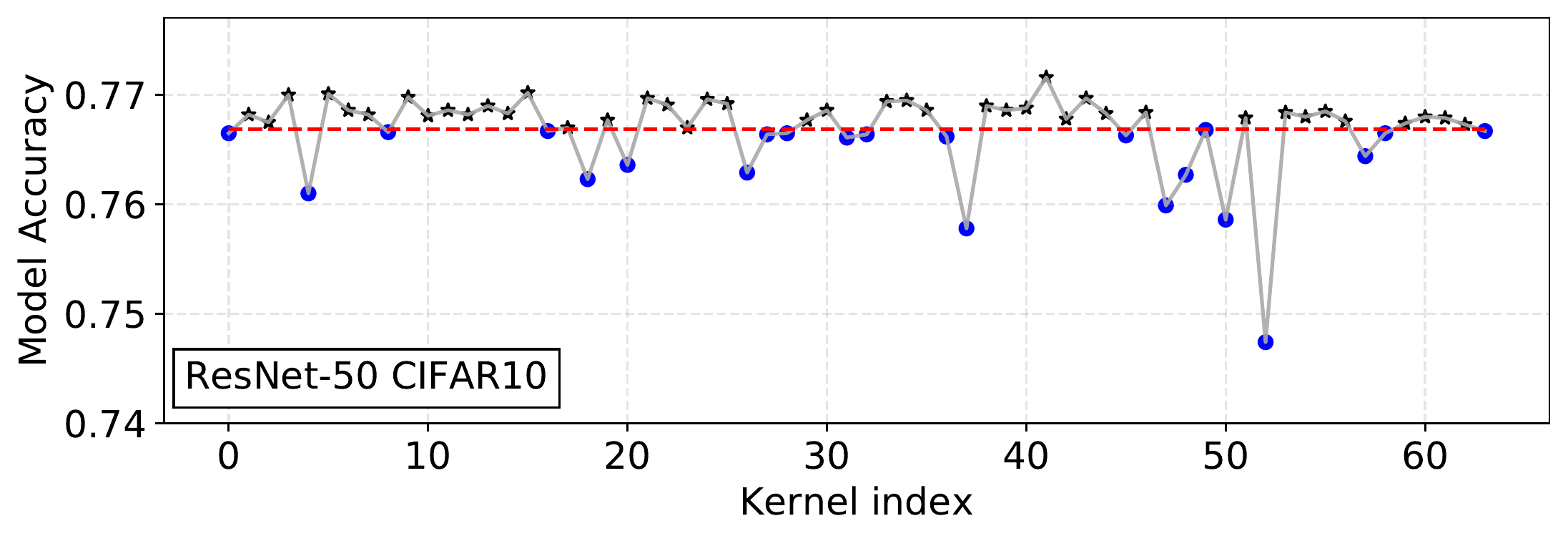}
    \caption{Evaluated ResNet-50 model trained for 10 epochs. Fragile kernels $S$ shown in blue below mean performance line in red and null kernels $S'$ are shown in black star above mean line in red.\label{fig:kernal_droput_exp}}
\end{figure}
\setlength\intextsep{0pt}
}
\section{Adversarial Targeting Algorithm}
We assuming that the parameters $\theta_{L,S'}$ of null kernels $S'$ in layer $L$, carry within them some noise that render the overall influence of these kernels on the model performance to be lower than the fragile kernel $S$, we propose to filter parameters $\theta_{L,S'}$ to remove noise. We assume the distribution of the noise in matrix $K_{L,\bar{S}}$ to be Gaussian noise. For this, we can use the works of Gavish and Donoho~\cite{gavish2014optimal} to recover a lower rank matrix from noisy data and retaining only the most important features. The filtering of $\theta_{L,S'}$ produces the modified parameters $\theta'_{L,S'}$. The filtered parameters relating to null kernels $\theta'_{L,S'}$, are said to be more robust if the probability of predicting the true class using modified model parameters $\theta'$ is higher than $\theta$ as per:
\begin{equation}
    P(\hat{y}=y|x',\theta') > P(\hat{y}=y|x',\theta).
\end{equation}

We compose a matrix $K_{L,S'}$ by stacking flattened null kernel parameters $\theta'_{L,S'}$ and compress $K_{L,S'}$ to remove noise or redundant information, thus increasing the influence of these null kernels $S'$ on the model's overall performance. While filtering the null kernels $S'$, we proportionally amplify fragile kernel $S$. This is to maintain relative magnitude of the local features within the network and propagate the essential representations to deeper layers of the network better.

\subsection{Filtering of Null Kernels $S'$}
\label{sec:filtering_null_kernels}
We decompose the null kernel's matrix $K_{L,S'}$ using singular value decomposition (SVD) and reduce the complexity of the representations by clamping all values below a filtering threshold $\tau$. We apply this method only to the first convolutional layer because of its susceptibility to any distortions having a higher influence on the network's performance~\cite{cheney2017robustness}. 
We use SVD to decompose our null kernel matrix $K_{L,\bar{S}}$ into its respective eigenvalues $\Sigma$ and eigenvectors matrices $U$ and $V$ as:
\begin{equation}
\label{eq:k_sbar}
    K_{L,S'} = U\Sigma V^T.
\end{equation}
We then compute a truncated matrix of singular values $\widetilde{\Sigma}$ by clamping all singular values to be at most equal to threshold value $\tau$ as per: 
\begin{equation}
    \widetilde{\sigma_i} = \arg \min(\sigma,\tau),
\end{equation}
where $\sigma$ is the diagonal of $\Sigma$ and $\widetilde{\sigma_i}$ is the row upto which the matrix $\sigma$ is truncated. 
The thresholding value $\tau$ for $m$-by$n$ matrix is given as:
\begin{equation}
\label{eq:threshold}
    \tau = \lambda(\beta)\cdot\sqrt{n}\varepsilon,
\end{equation}
where $\beta=m/n$, $\varepsilon$ is the noise level within the matrix, and the term $\lambda(\beta)$ is expressed as~\cite{gavish2014optimal}:
\begin{equation}
    \label{eq:lambda_beta}
    \lambda(\beta) = \sqrt{2(\beta+1)+\frac{c_1\beta}{(\beta+1)+\sqrt{\beta^2+c_2\beta+1}}},
\end{equation}
where constants $c_1$ and $c_2$ respectively are 8 and 14.

We then find the noise level value $\varepsilon$ in~\eqref{eq:threshold} experimentally through a systematic search method using a sample set of the parameters. As the final filtering  step, we reconstruct the filtered weight matrix $\widetilde{K}_{L,\bar{S}}$ by using the clamped singular values and corresponding eigenvectors as: 
\begin{equation}
\label{eq:reconstruct_sbar}
    \widetilde{K}_{L,\bar{S}} = U\widetilde{\Sigma}V^T.
\end{equation}

\subsection{Amplification of Fragile Kernels $S$}
\label{sec:amplifying_s}

The amplification of fragile kernels parameters matrix $K_{L,S}$ by a scaling factor of $\alpha$ computed using~\eqref{eq:k_sbar} and~\eqref{eq:reconstruct_sbar} as per:
\begin{equation}
    \widetilde{K}_{L,S} = \alpha K_{L,S},
\end{equation}
where scaling factor of $\alpha$ is
\begin{equation}
    \alpha = 1+||K_{L,\bar{S}}-\widetilde{K}_{L,\bar{S}}||_{2}.
\end{equation}
The aim of this process is to amplify the features within fragile kernels $S$, such that a greater magnitude of adversarial perturbation is required to vary such kernels. 

\subsection{Adversarial Targeting of Fragile and Null Kernels}
\label{sec:adversarial_targeting}
We assess the robustness of the fragile kernels $S$ and null kernels $S'$ by our robustness targeting algorithm shown in Algorithm~\ref{algo:robustness_trageting}. The FSGM attack for varied range of perturbations $\epsilon$ is used to compute the evaluated first convolutional layer's outputs $\hat{y}_{x}$ and $\hat{y}_{x'}$. The mean difference between each kernel in the output of $\hat{y}_{x}$ and $\hat{y}_{x'}$ are calculated and compared to see which is highest, indicating a greater average concentration of the attack.

\begin{algorithm}[ht]
	\caption{Adversarial targeting\label{algo:robustness_trageting}} 
	\begin{algorithmic}[1]
	\State Initialise $f() \rightarrow f_L()$ \Comment{$f_{L}()$ is the $L$-th layer of full network $f()$ }
	\State Compute indices of fragile kernels $S$ and null kernels $S'$ as per Sec~\ref{sec:frigile_and_null_kernerls}
	\State $S_{attack} = \{ \}$ \Comment{an empty list to store examples that attacks $S$}
		\For {perturbation $\epsilon \in \mathbb{R}$} \Comment{ where $\epsilon$ is perturbation magnitude}
		    \State \textit{attack} = FGSM($f_{L}, \epsilon)$
		    \State $S_{\text{count}} = 0$ 

			\For {($x, y$) in ($X_{test}, Y_{test}$)}
				\State $x' = \textit{attack}(x, y)$ \Comment{create an adversarial example $x'$ for input $x$ and level $y$} 
				\State $\hat{y}_{x} = f_{L}(x)$ \Comment{output of $L$-th layer on unperturbed input $x$}
				\State $\hat{y}_{x'} = f_{L}(x')$ \Comment{output of $L$-th layer on perturbed input $x'$}
				\State $\textbf{d} = ||\hat{y}_{x} -  \hat{y}_{x'} ||_2$ \Comment{Euclidean distance $\textbf{d} = (d_1, \ldots, d_{k})$ between $\hat{y}_{x}$ and $\hat{y}_{x'}$}
				\State $S_f = (\sum_{j}^{|S|} d_{j,S} )/ |S| $ \Comment{Average of distances $d_{j,S}$ of all $S$ select from $\textbf{d}$}
				\State $S_n = (\sum_{j}^{|S'|} d_{j,S'})/ |S'| $ \Comment{Average of distances $d_{j,S'}$ of all $S'$ select from $\textbf{d}$}
				\If{$S_f > S_n$}
				    \State $S_{\text{count}} = S_{\text{count}} + 1$  \Comment{increase counter of attacks for fragile kernels}
				\EndIf
			\EndFor
			\State $S_{attack} \leftarrow S_{\text{count}}$ \Comment{add $S_{\text{count}}$ to the list $S_{attack}$}
		\EndFor
	\end{algorithmic} 
\end{algorithm}



\section{Results and Discussion}
\begin{figure}[th]
    \centering
    \includegraphics[width=1\linewidth]{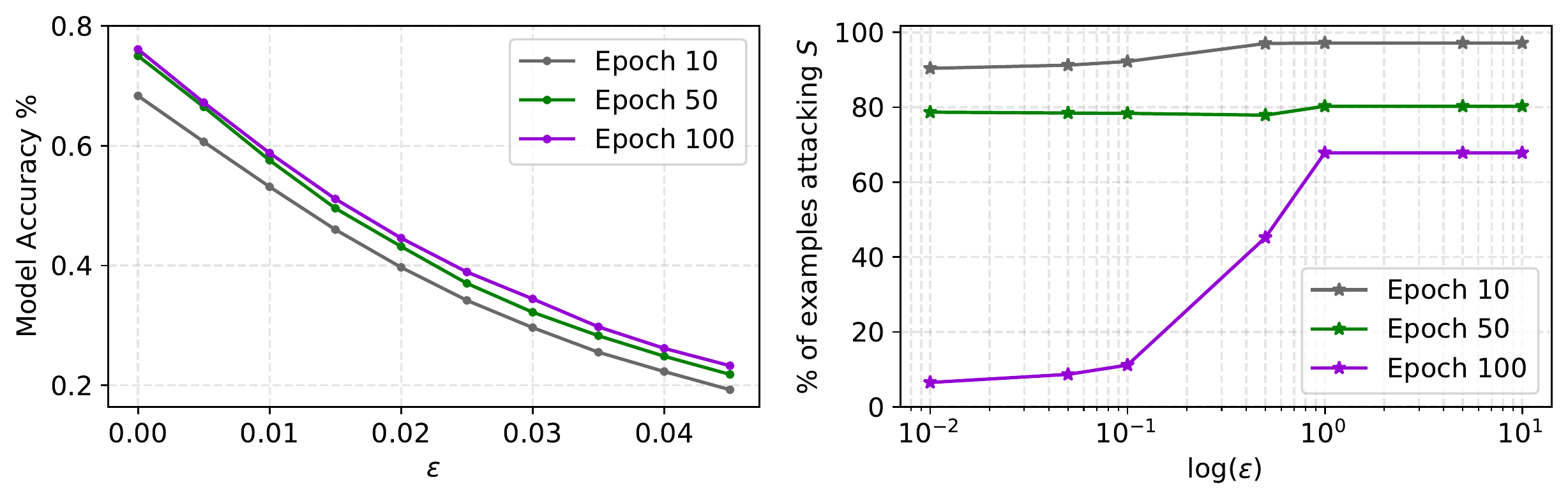}
    \caption{\textit{Left:} ResNet-50 model trained on the CIFAR10 dataset for epochs 10, 50 and 100 against the FGSM attack, with $\epsilon$ increasing linearly, marked by dots. \textit{Right:} ResNet-50 model trained on the CIFAR10 dataset for epochs 10, 50 and 100 against the FGSM attack, with attack magnitude increasing logarithmically, marked by star symbols.  Epoch 10, 50, 100 respectively indicated in colors grey, green, violet.}
    \label{fig:targetting_and_robustness}
\end{figure}

\begin{figure}
    \centering
    \includegraphics[width=1\linewidth]{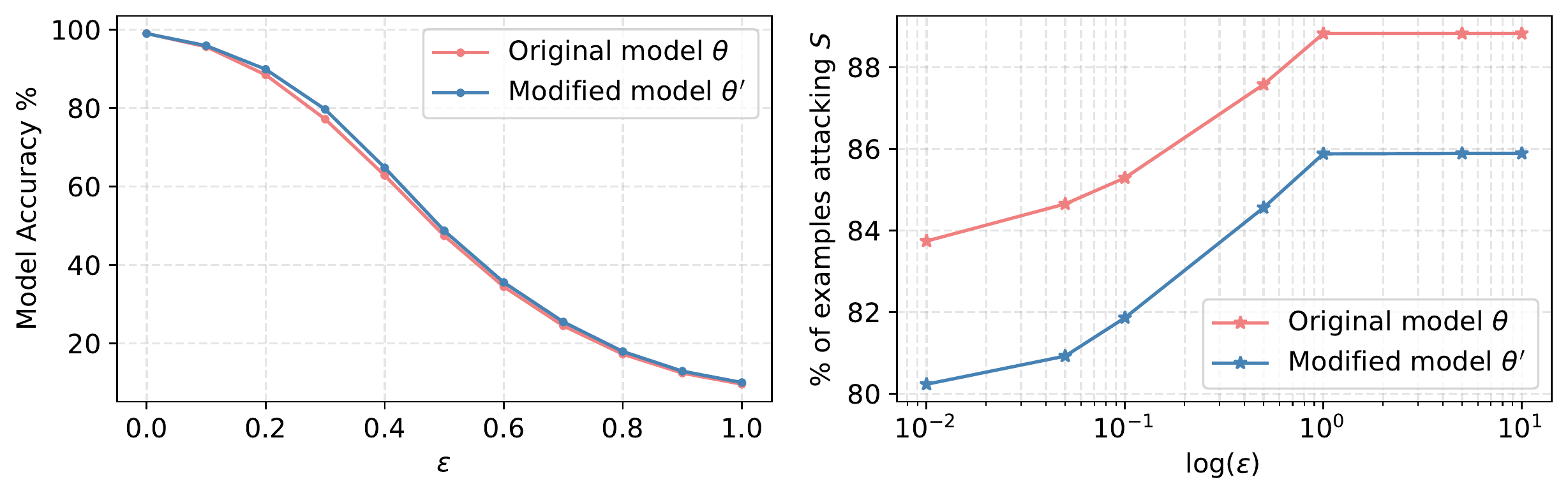}
    \caption{Concentration of the adversarial attack on fragile kernels $S$ for both the original model with parameters $\theta_L$ and the modified model, with $\theta'_L$ in a ResNet-50 model trained on the MNIST dataset for 10 epochs, using the methods proposed in Sections.~\ref{sec:filtering_null_kernels} and~\ref{sec:amplifying_s}.}
    \label{fig:adv_target_50_100}
\end{figure}

In first series of experiments, we use the two sets $S$ and $S'$ obtained as per Fig.~\ref{fig:kernal_droput_exp} on the ResNet-50 model and apply them to Algorithm~\ref{algo:robustness_trageting} using the CIFAR10 dataset, resulting in Fig.~\ref{fig:targetting_and_robustness} and the MNIST dataset, resulting in Fig.~\ref{fig:adv_target_50_100}: 

For Fig.~\ref{fig:targetting_and_robustness}, we measure the robustness of ResNet-50 models and compare the percentage of examples attacking fragile kernels $S$ and the model accuracy against FGSM attack. In  Fig.~\ref{fig:targetting_and_robustness} (\textit{Left}), we notice that as the number of training epochs increases, the model's accuracy also increases for both the unperturbed ($\epsilon=0$) and perturbed ($\epsilon>0$) examples. In  Fig.~\ref{fig:targetting_and_robustness} (\textit{Right}), using the results from the adversarial targeting Algorithm~\ref{algo:robustness_trageting}, we also notice that the percentage of examples attacking fragile kernels $S$ is higher for highly perturbed examples. However, for smaller perturbation magnitudes, 100 epoch model is more robust. This suggests that as the model becomes more robust (from epoch 10 to 100), the percentage of examples attacking fragile kernels $S$ and null kernels $S'$ tends to distribute equally.

After applying our framework proposed in sections~\ref{sec:filtering_null_kernels} and~\ref{sec:amplifying_s} using $\varepsilon$ value of $0.015$ to the first convolutional layer $\theta_{L}$, resulting in filtered layer parameters $\theta'_{L}$, we observe the difference in attack distribution between original model and modified model using Algorithm~\ref{algo:robustness_trageting}. 

We apply the parameter filtering framework to a ResNet-50 model trained on the MNIST for 10 epochs. The results of which is shown in Fig.~\ref{fig:adv_target_50_100}. In this experiment, although the number of fragile kernels $S$ are $37\%$ of the total kernels within the layer, these kernels show a larger average distance between the outputs of the original layer $\theta_L$ and modified layer $\theta'_L$ for almost $89\%$ of the tested input examples on original model. Furthermore, as the attack strength is increased by increasing $\epsilon$, the average magnitude of the attack on kernels $S$ also increased. However, our method of filtering parameters $\theta'_L$ kept the percentage of tested examples attacking fragile kernels $S$ lower than the original model.

\begin{figure}[!ht]
    \centering
\begin{minipage}{.5\textwidth}
  \centering
  \includegraphics[width=1\linewidth]{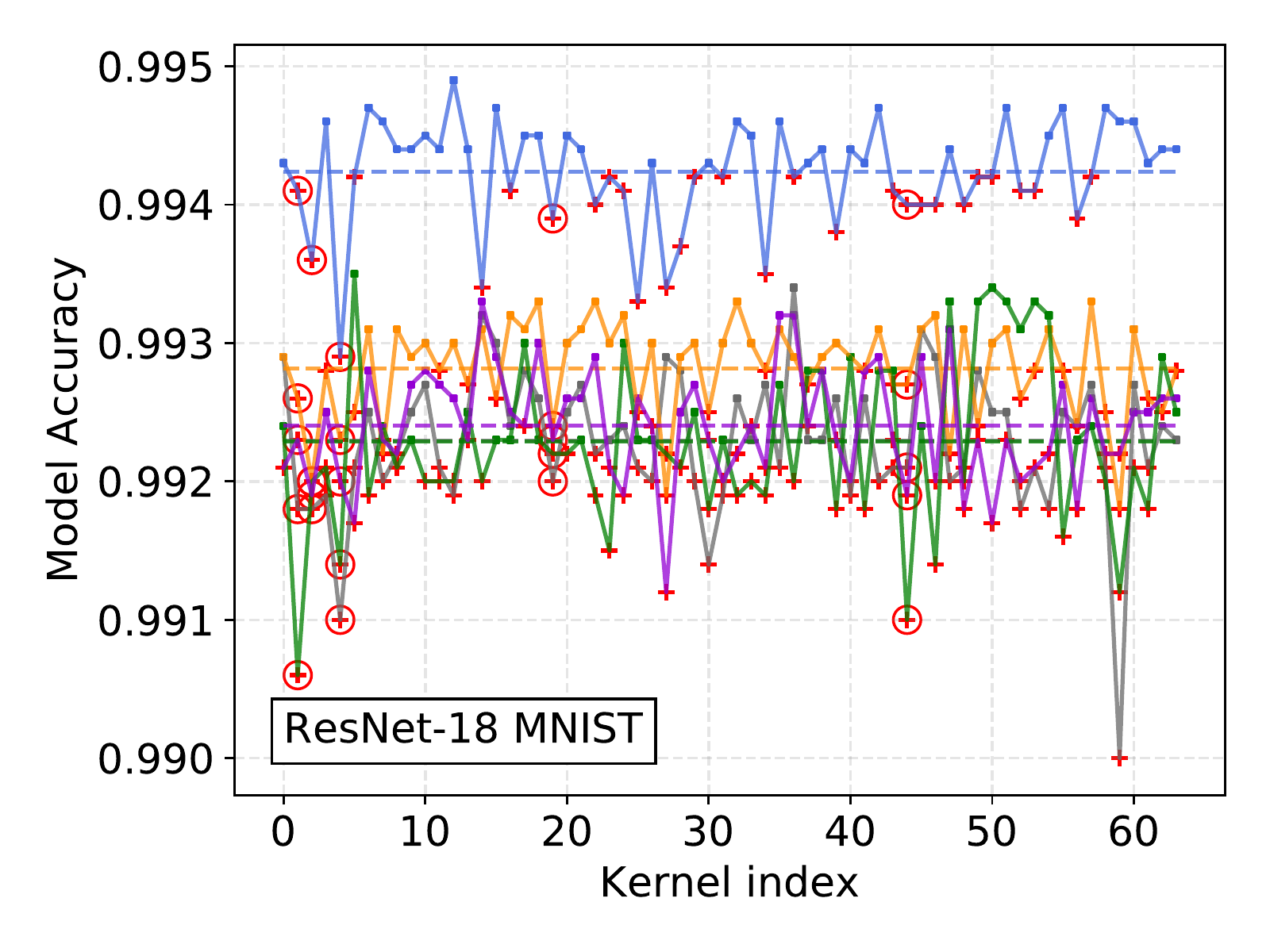}
\end{minipage}%
\begin{minipage}{.5\textwidth}
  \centering
  \includegraphics[width=1\linewidth]{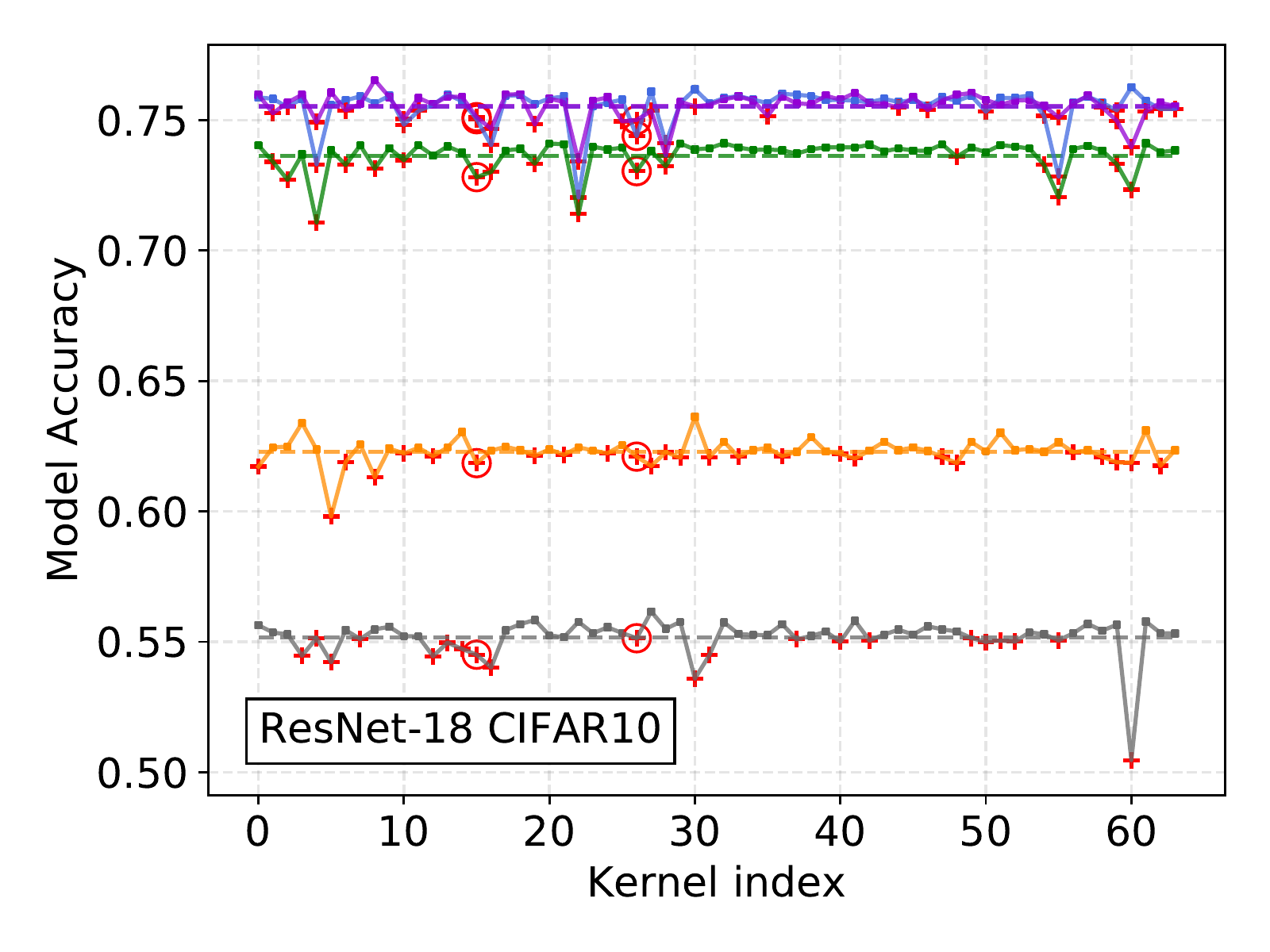}
\end{minipage}

\begin{minipage}{.5\textwidth}
  \centering
  \includegraphics[width=1\linewidth]{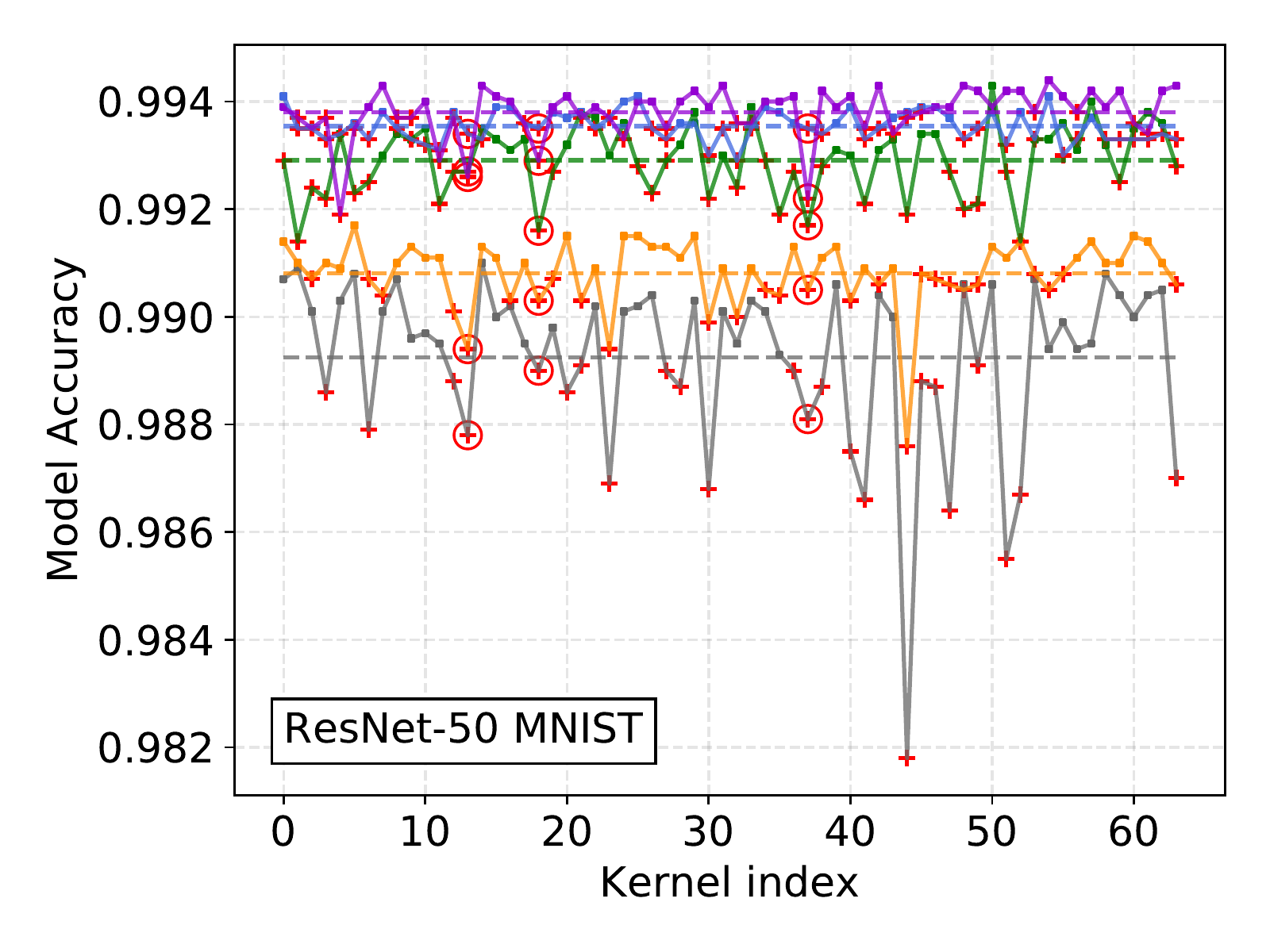}
\end{minipage}%
\begin{minipage}{.5\textwidth}
  \centering
  \includegraphics[width=1\linewidth]{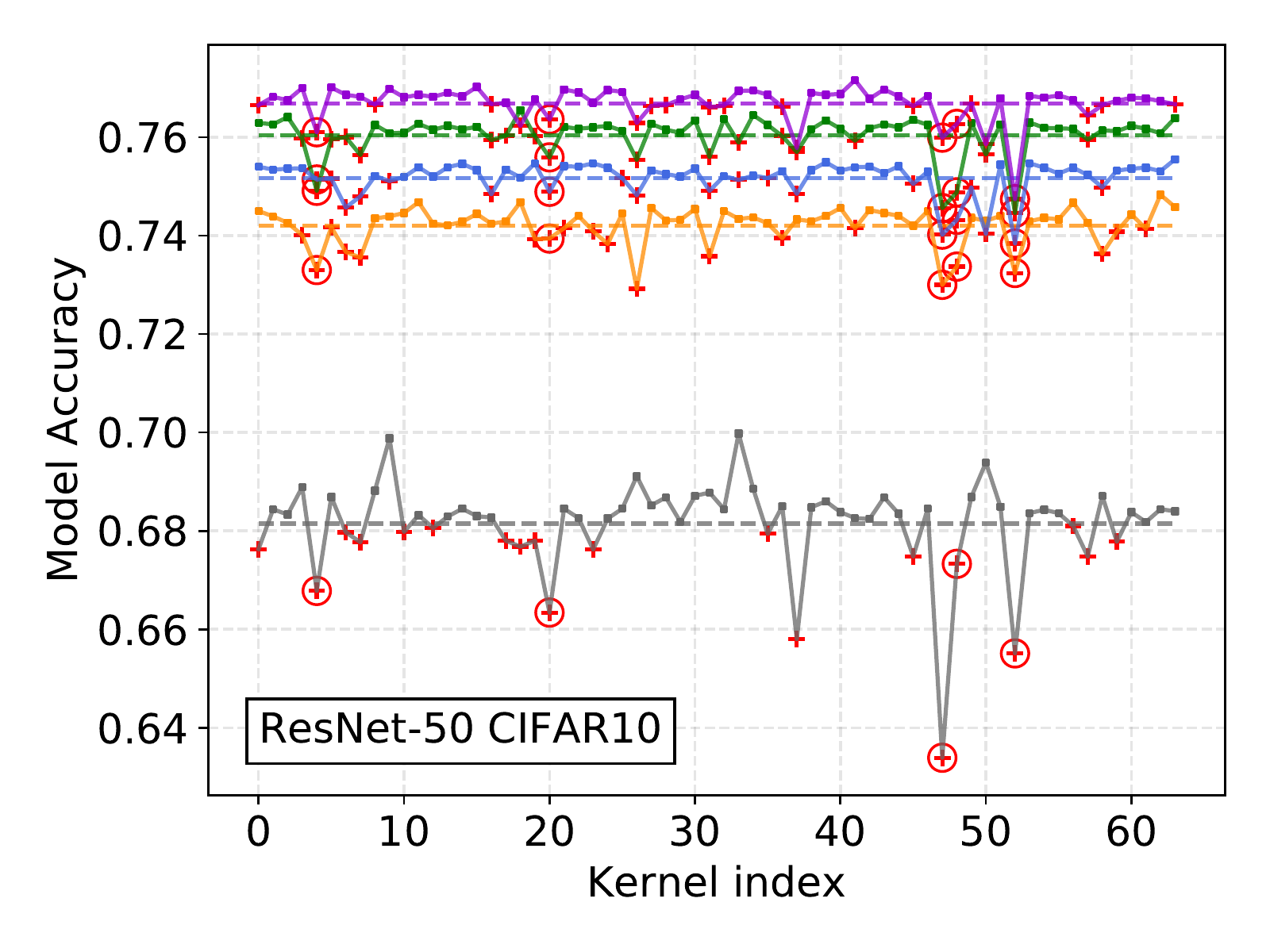}
\end{minipage}

\begin{minipage}{.5\textwidth}
  \centering
  \includegraphics[width=1\linewidth]{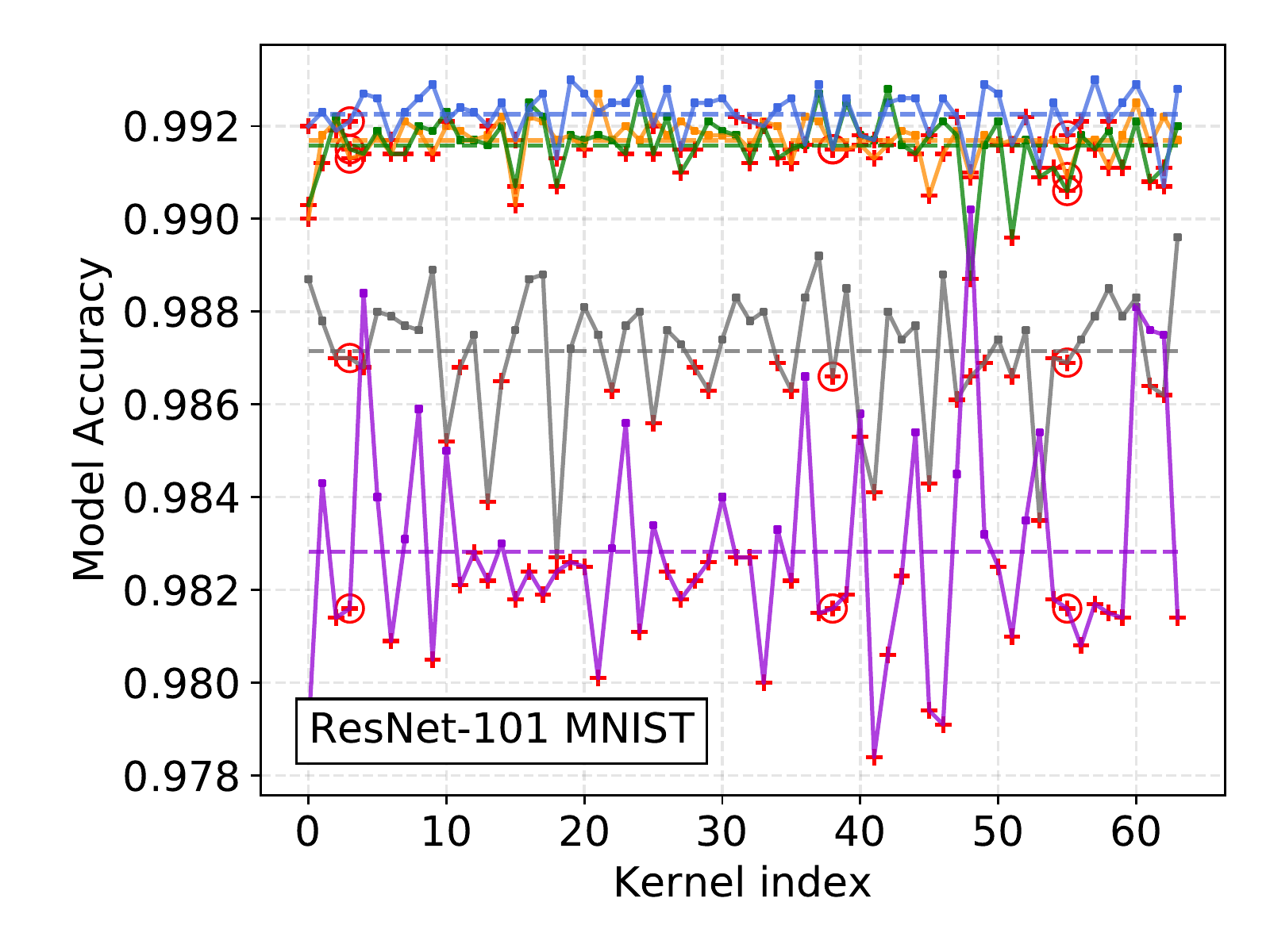}
\end{minipage}%
\begin{minipage}{0.5\textwidth}
  \centering
  \includegraphics[width=1\linewidth]{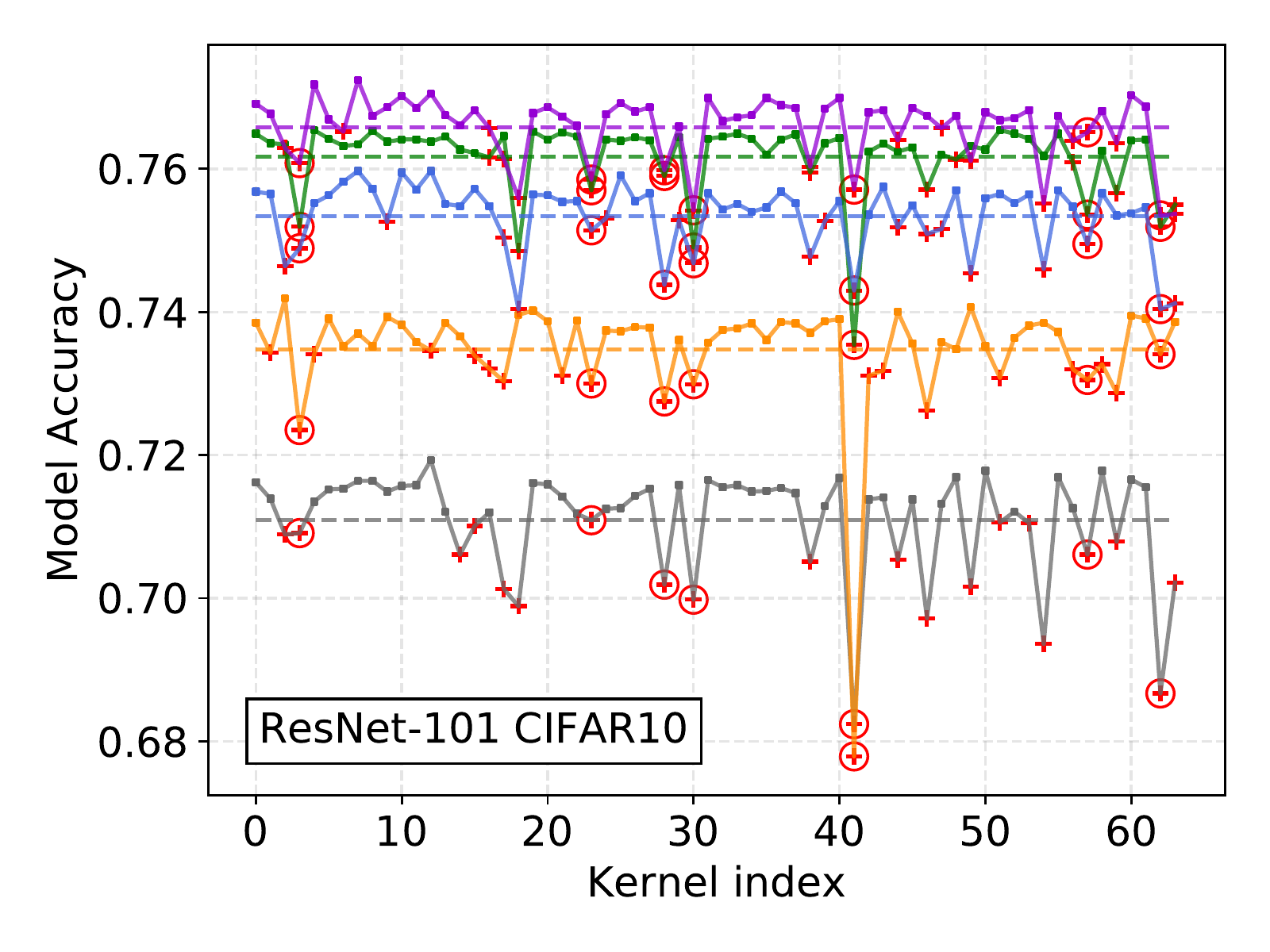}
\end{minipage}

\centering
\includegraphics[width=1\linewidth]{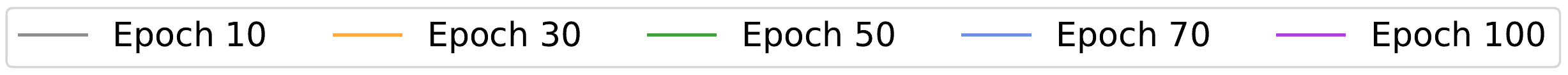}

\caption{Variance of model performance to individual kernels being dropped out within the first convolutional layer. Red circles indicate fragile kernels that remain fragile throughout the all training epochs, whereas red crosses indicate kernels that are observed as fragile for the specific training epoch length.} 

\label{fig:resnet_dropout_training}

\end{figure}

We observe from 
Fig.~\ref{fig:resnet_dropout_training} how the influence of kernels in the first convolutional layer varies during the training process while we systematically drop and assess the kernels. In Fig.~\ref{fig:resnet_dropout_training}, red circles are the kernels that carry a higher influence through all stages of model training. We notice that as we change the model from ResNet-18 to ResNet-50 and ResNet-101, the number of influential fragile kernels increases on the CIFAR10 dataset. This is as we may expect, model architectures with greater complexities are able to learn the important features from the dataset faster than shallower model architectures. We notice from Fig.~\ref{fig:resnet_dropout_training}, that the average model performance of the kernels in $\theta_{L}$ increases to a limit for models trained on the CIFAR10 dataset and shows to increase and then decrease for the models trained on the MNIST dataset. This characteristic invites a separate set of experiments to better understand how model overfitting affects nodal dropouts.

\section{Conclusion}
\label{sec:con}
In this study we show how an FGSM attack targets specific neurons within the first convolutional layer of ResNet-18, ResNet-50 and ResNet-101 models trained on both the CIFAR10 and NNIST datasets. To prove this property, we first identify fragile kernels $S$ and null kernels $S'$ sets within the evaluated layer using an iterative dropout method and measuring the variance in model performance. We use the kernel indices of $S$ and $S'$ to evaluate the highest average distance between the outputs of the layer using the original input $x$ and perturbed example $x'$. In doing so, we find that for a ResNet-50 model trained on the CIFAR10 dataset for 50 epochs, the number of fragile kernels $S$ account to $37\%$ of the total number of kernels in the layer yet show to have a higher average difference for approximately $89\%$ of the examples evaluated.

We also show how the robustness against the FGSM attack, and the targeting of fragile kernels $S$ varies as the model is trained, thus showing a correlation between a model becoming more robust and the targeting of fragile kernels. Furthermore, we propose a layer parameter filtering algorithm that improves robustness in a model by removing information from null kernels $S'$ and amplifying the information in $S$. This simple method, despite only being applied to the first convolutional layer, improves the robustness of a model with less training. It should be noted that, although our study focuses on the first convolutional layer only due to the layer being highly influence over the model's performance, other layers can also be evaluated using this proposed framework.

%
%
\bibliographystyle{splncs04}
\bibliography{mybibliography.bib}





\end{document}